\documentclass{article} % For LaTeX2e

\usepackage{amsmath,amsfonts,bm}

\def\eqref#1{equation~\ref{#1}}
\def\1{\bm{1}}

\DeclareMathAlphabet{\mathsfit}{\encodingdefault}{\sfdefault}{m}{sl}
\SetMathAlphabet{\mathsfit}{bold}{\encodingdefault}{\sfdefault}{bx}{n}

\newcommand{\E}{\mathbb{E}}

\DeclareMathOperator*{\argmin}{argmin}

\usepackage{amsmath}
\usepackage{amssymb}
\usepackage{booktabs}
\usepackage{graphicx}
\usepackage{makecell}
\usepackage{microtype}
\usepackage{multirow}
\usepackage{paralist}
\usepackage{url}
\usepackage{xcolor}
\usepackage{xspace}

\usepackage{algorithm2e}

\usepackage[accepted]{icml2020}

\usepackage[hyperfootnotes=false]{hyperref}   %%% REMOVE [draft] BEFORE SUBMISSION!

\newcommand{\papername}{\textsc{AFLite}\xspace}
\newcommand{\afopt}{\textsc{AFOpt}\xspace}

\newcommand{\roberta}{$\textit{RoBERTa}$\xspace}
\newcommand{\robphi}{$\phi_{\textit{RoBERTa}}$\xspace}
\newcommand{\robfilt}{$D(\phi_{\textit{RoBERTa}})$\xspace}

\newcommand{\bert}{$\textit{BERT}$\xspace}
\newcommand{\bertphi}{$\phi_{\textit{BERT}}$\xspace}
\newcommand{\bertfilt}{$D(\phi_{\textit{BERT}})$\xspace}

\newcommand{\glove}{$\textit{GloVe}$\xspace}

\newcommand{\hyp}{\small \textit{-HypOnly}}
\newcommand{\partialinput}{\small \textit{-PartialInput}}

\begin{document}
\twocolumn[
\icmltitle{Adversarial Filters of Dataset Biases}

% It is OKAY to include author information, even for blind
% submissions: the style file will automatically remove it for you
% unless you've provided the [accepted] option to the icml2019
% package.

% List of affiliations: The first argument should be a (short)
% identifier you will use later to specify author affiliations
% Academic affiliations should list Department, University, City, Region, Country
% Industry affiliations should list Company, City, Region, Country

% You can specify symbols, otherwise they are numbered in order.
% Ideally, you should not use this facility. Affiliations will be numbered
% in order of appearance and this is the preferred way.
\icmlsetsymbol{equal}{*}

\begin{icmlauthorlist}
\icmlauthor{Ronan {Le Bras}}{ai2}
\icmlauthor{Swabha Swayamdipta}{ai2}
\icmlauthor{Chandra Bhagavatula}{ai2}
\icmlauthor{Rowan Zellers}{ai2,uw}
\icmlauthor{Matthew E. Peters}{ai2}
\icmlauthor{Ashish Sabharwal}{ai2}
\icmlauthor{Yejin Choi}{ai2,uw}
\end{icmlauthorlist}

%\icmlaffiliation{ai2}{Allen Institute for AI}
\icmlaffiliation{ai2}{Allen Institute for Artificial Intelligence}
\icmlaffiliation{uw}{Paul G. Allen School of Computer Science, University of Washington}

\icmlcorrespondingauthor{Ronan Le Bras, Swabha Swayamdipta, Chandra Bhagavatula}{\{ronanlb,swabhas,chandrab\}@allenai.org}

% You may provide any keywords that you
% find helpful for describing your paper; these are used to populate
% the "keywords" metadata in the PDF but will not be shown in the document
\icmlkeywords{Machine Learning, ICML}

\vskip 0.3in
]

\printAffiliationsAndNotice{}
% \maketitle
\setcounter{footnote}{2}

\begin{abstract}

% Revised abstract to comply with ICML guidelines:  
% "Abstracts must be a single paragraph, ideally between 4–6 sentences long. Gross violations will trigger corrections at the camera-ready phase."
Large neural models have demonstrated human-level performance on language and vision benchmarks, while their performance degrades considerably on adversarial or out-of-distribution samples.
This raises the question of whether these models have learned to solve a \emph{dataset} rather than the underlying \emph{task} by overfitting to \textit{spurious dataset biases}. 
We investigate one recently proposed approach, \papername, which adversarially filters such dataset biases, as a means to mitigate the prevalent overestimation of machine performance. 
We provide a theoretical understanding for \papername, by situating it in the generalized framework for optimum bias reduction.
%
%Our experiments show that as a result of the substantial reduction of these biases, models trained on the filtered datasets yield better generalization to out-of-distribution tasks, especially when the benchmarks used for training are over-populated with biased samples.
%We show that \papername is broadly applicable to a variety of both real and synthetic datasets for reduction of measurable dataset biases and provide extensive supporting analyses.
We present extensive supporting evidence that \papername is broadly applicable for reduction of measurable dataset biases, and that models trained on the filtered datasets yield better generalization to out-of-distribution tasks.
Finally, filtering results in a large drop in model performance (e.g., from 92\% to 62\% for SNLI), while human performance still remains high. 
Our work thus shows that such filtered datasets can pose new research challenges for robust generalization by serving as upgraded benchmarks.

\end{abstract}

\section{Introduction}
\label{sec:intro}

\begin{figure*}[ht]
    \centering
    \vspace*{12pt}
    \hspace*{-0.25in} \includegraphics[width=1.04\textwidth]{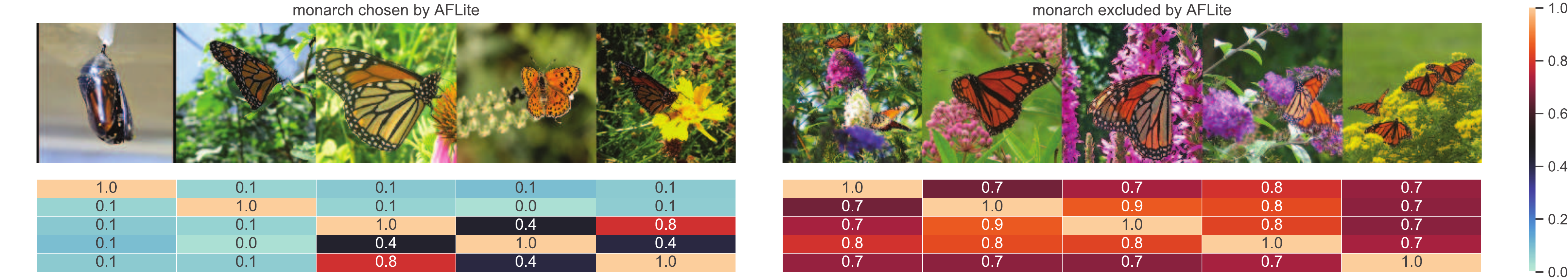}\\
    \vspace*{12pt}
    \hspace*{-0.25in} \includegraphics[width=1.04\textwidth]{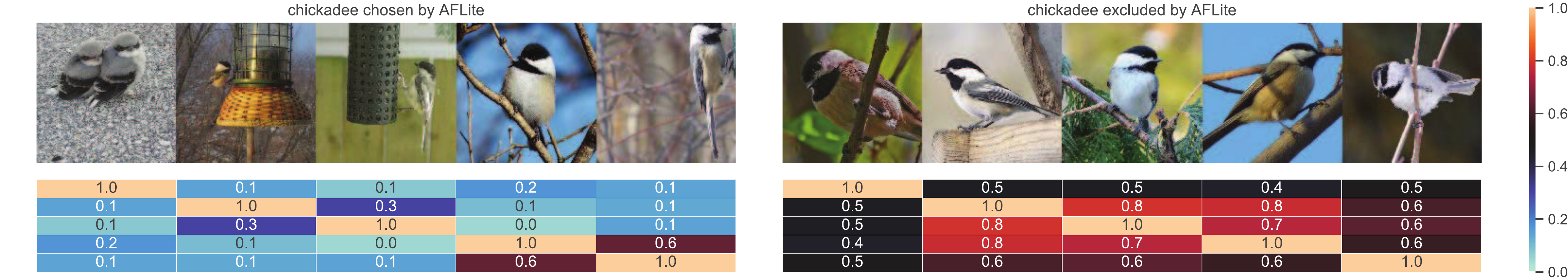}
    \caption{Example images of the Monarch Butterfly and Chickadee from ImageNet. 
    On the left are images in each category which were retained (filtered in) by \papername{}, and on the right, the ones which were removed (filtered out). 
    The heatmap shows pairwise cosine similarity between EfficientNet-B7 features \citep{tan2019efficientnet}. 
    The retained images (left) show significantly greater diversity -- such as the cocoon of a butterfly, or the non-canonical chickadee poses -- also reflected by the cosine similarity values. 
    This diversity suggests that the \papername{}-filtered examples present a more accurate benchmark for the task of image classification. }
    % , as opposed to fitting to particular dataset biases.}
    \label{fig:imagenetexamples}
\end{figure*}

%\ronanlb{Todo list for near camera-ready pdf, due on Jun 29 AOE:\\
%x Acknowledgements\\ - Proofreading\\ - Integrating reviewers' comments}

Large-scale neural networks have achieved superhuman performance across many popular AI benchmarks, for tasks as diverse as image recognition (ImageNet; \citealp{Russakovsky2015Imagenet}), natural language inference (SNLI; \citealp{Bowman2015ALA}), and question answering (SQuAD; \citealp{Rajpurkar2016SQuAD10}).
However, the performance of such neural models degrades considerably when tested on out-of-distribution or adversarial samples, otherwise known as data ``in the wild'' \citep{Eykholt2018RobustPA,Jia2017AdversarialEF}.
% when taken out of these dataset environments and evaluated on %non-identically distributed
This phenomenon indicates that high performance of the strongest AI models is often confined to specific \textit{datasets}, implicitly making a closed-world assumption.
In contrast, true learning of a \textit{task} necessitates generalization, or an  open-world assumption. %, which are far from being solved.
A major impediment to generalization is the presence of spurious \emph{biases} -- unintended correlations between input and output -- in existing datasets \citep{Torralba2011UnbiasedLA}. 
% \ashish{the phrase `spurious biases between input and output' isn't quite clear (`bias between X and Y'). Perhaps something like `... spurious biases, i.e, uninteresting/unintended correlations between input and output'?} 
Such biases or \emph{artifacts}\footnote{We will henceforth use \emph{biases} and \emph{artifacts} interchangeably.} 
are often introduced during data collection \citep{fouhey2018lifestyle} or during human annotation \citep{Rudinger2017SocialBI,gururangan-etal-2018-annotation,poliak-etal-2018-hypothesis,tsuchiya-2018-performance,Geva2019AreWM}. 
Not only do dataset biases inevitably bias the models trained on them,
% Evidence from recent work points in the direction of the ease of obtaining and labeling much of the data, which is not necessarily representative of the task we seek to measure. \cscomment{I don't understand the prev sentence.}
but they have also been shown to significantly inflate model performance, leading to an overestimation of the true capabilities of current AI systems \cite{sakaguchi2019winogrande,hendrycks2019natural}.
% \yejin{cite winogrande and...?}\ashish{could cite ImageNet-A... same message}.

% Answering this question is key because benchmarks serve important roles in the community. 
% Not only do they \emph{direct} progress on core tasks, they also make it easier to tackle the lofty target tasks such as image recognition in the wild through a more practically https://www.overleaf.com/project/5d66bab43845ab1394890c82-scoped dataset such as ImageNet. 

%\swabha{TODO: Expand on this by touching on related work.} 
% Many recent studies have addressed this concern by proposing approaches to reduce task- or dataset-specific biases
Many recent studies have investigated task or dataset specific biases, including language bias in Visual Question Answering \citep{goyal2017making}, texture bias in ImageNet \citep{geirhos2018imagenet}, and hypothesis-only reliance in Natural Language Inference \citep{gururangan-etal-2018-annotation}. These studies have yielded domain-specific algorithms to address the found biases.
% \yejin{(... brief descriptions (such as gender bias? sentiment bias??) ...) to highlight how limited scopes of biases previous approaches have addressed} \citep{goyal2017making, geirhos2018imagenet,He2019UnlearnDB,Li2019REPAIRRR}.
However, the vast majority of these studies follow a \emph{top-down} framework where the bias reduction algorithms are essentially guided by researchers' intuitions and domain insights on particular types of spurious biases. 
While promising, such approaches are fundamentally limited by what the algorithm designers can \textit{manually} recognize and enumerate as unwanted biases. 

% In this work, we propose to reduce biases on \emph{any benchmark dataset} in a \emph{bottom-up} approach to algorithmic bias reduction. 
Our work investigates \papername{}, an alternative \emph{bottom-up} approach to \textit{algorithmic} bias reduction. 
\papername{}\footnote{Stands for Lightweight Adversarial Filtering.} was recently proposed by \citet{sakaguchi2019winogrande}---albeit very succinctly---to systematically discover and filter \emph{any} dataset artifact in crowdsourced commonsense problems.
\papername{} employs a model-based approach with the goal of removing spurious artifacts in data beyond what humans can intuitively recognize, but those which are exploited by powerful models. 
% Since it removes spurious data that models tend to rely on, it is adversarial to any model.
Figure \ref{fig:imagenetexamples} illustrates how \papername{} reduces dataset biases in the ImageNet dataset for object classification. 
% The original ImageNet (shown on the right column) \swabha{incorrect, both are original imagenet!} demonstrates stronger \emph{camera bias} and \emph{search engine bias}, extremely skewing samples toward many similar prototypical samples.
% \swabha{worried about the above terms, are these well known in the vision community? if so are there any citations? if not let's remove.}
% This drastically differs from the true distribution of images in the wild \cite{Beery2018Recognition}. 
% \papername{} essentially flattens such skewed sample distributions so that the dataset better represents the diverse samples of each category, as shown in the left column of Figure \ref{fig:imagenetexamples}.
% \swabha{Also, we have a similar discussion in the caption, do we need to also include it in the intro?}

%However, the complex artifacts that emerge from large-scale dataset creation are challenging to exhaustively identify and remove via optimization techniques or model architectures, since biased and unbiased features are too intertwined in making a model prediction.
%While it has been used for creating several datasets, such as Winogrande \cite{sakaguchi2019winogrande}, Physical-IQA \todo{\cite{}} and Social IQA \todo{\cite{}}, there have been no efforts towards a general theoretical understanding and broader applicability of the approach.

This paper presents the first theoretical understanding and comprehensive empirical investigations into \papername{}. 
More concretely, we make the following four novel contributions.

First, we situate \papername{} in a theoretical framework for optimal bias reduction, and demonstrate that \papername provides a practical approximation of \afopt, the ideal but computationally intractable bias reduction method under this framework (\S\ref{sec:algo}). 
% First, we situate \papername{} in a theoretical framework for optimal bias reduction (\afopt{} \ashish{odd to view \afopt{} as a framework; is it a method that achieves optimum reduction under this framework?}), demonstrating that \papername provides a tractable approximation \ashish{of what?} (\S\ref{sec:algo}). \ashish{one rephrasing that may address both items: we situate ... reduction, and demonstrate that \papername provides a practical approximation of \afopt, the ideal but computationally intractable bias reduction method under this framework.} \ashish{\textsc{AFOpt} instead of \afopt?}

Second, we present an extensive suite of experiments that were lacking in the work of \citet{sakaguchi2019winogrande}, to validate whether \papername{} truly removes spurious biases in data as originally assumed.
Our baselines and thorough analyses use both synthetic (thus easier to control) datasets (\S\ref{sec:exp_syn}) as well as real datasets.
% in language and vision.
The latter span benchmarks across NLP (\S\ref{sec:exp_nlp}) and vision (\S\ref{sec:exp_vision}) tasks: the SNLI \citep{Bowman2015ALA} and MultiNLI \citep{williams-etal-2018-broad} datasets for natural language inference, QNLI \citep{Wang2018GLUEAM} for question answering, and the ImageNet dataset \citep{Russakovsky2015Imagenet} for object recognition.

Third, we demonstrate 
% \ashish{drop `surprising'? could just say: `We demonstrate that models trained on ... generalize substantially (??) better than...'. }
that models trained on \papername{}-filtered data generalize substantially better to out-of-domain samples, compared to models that are trained on the original biased datasets (\S\ref{sec:exp_nlp}, \S\ref{sec:exp_vision}). 
% For instance, on a diagnostic evaluation for NLI, a model trained on the \papername{}-filtered portion of the SNLI dataset improves performance by as high as 80\% (\textit{constituent heuristic subset} \cite{mccoy-etal-2019-right}) compared to a model trained on the full dataset, in spite of containing far fewer training samples. 
% \ashish{a glimpse of the results would be valuable. Can we quantitatively highlight one strong generalization result as an example?}
These findings indicate that spurious biases in datasets make benchmarks artificially easier, as models learn to overly rely on these biases instead of learning more transferable features, thereby hurting out-of-domain generalization.
% \swabha{Include next sentence if we do Winogrande:
% In contrast to other approaches such as \citet{??}, we show that this method is agnostic to the nature of the bias (e.g. gender bias, or partial input bias).}

Finally, we show that \papername{}-filtering makes widely used AI benchmarks considerably more challenging.
We consistently observe a significant drop in the in-domain performance even for state-of-the-art models on all benchmarks, even though human performance still remains high; this suggests that currently reported performance on benchmarks might be inflated.
For instance, the best model on SNLI-\papername achieves only 63\% accuracy, a 30\% drop compared to its accuracy on the original SNLI.
These findings are especially surprising since \papername{} maintains an identical train-test distribution, while retaining a sizable training set.
% makes the benchmarks considerably harder for state-of-the-art models when tested for in-domain samples, 

In summary, \papername{}-filtered datasets can serve as upgraded benchmarks, posing new research challenges for robust generalization.
\section{\papername{}}
\label{sec:algo}

\setlength{\abovedisplayskip}{3pt}

Large datasets run the risk of prioritizing performance on the data-rich \emph{head} of the distribution, where examples are plentiful, and discounting the \emph{tail}. 
% Not sure we need to cite WinoGrande again here \papername{}~\changedAshish{\cite{sakaguchi2019winogrande}} 
\papername{} seeks to minimize the ability of a model to exploit biases in the head of the distribution, while preserving the inherent complexity of the \emph{tail}. 
% In this section, we provide a formal motivation for \papername{}, leading up to the algorithm.\footnote{While the algorithm was first proposed in \citet{sakaguchi2019winogrande}, we are the first to provide a supporting theoretical framework.}
In this section, we provide a formal framework for studying such bias reduction techniques, revealing that \papername{} can be viewed as a practical approximation of a desirable but computationally intractable optimum bias reduction objective.

\paragraph{Formalization} 

Let $\Phi$ be any feature representation defined over a dataset $\mathcal{D}=(X,Y)$. 
\papername{} seeks a subset $S \subset \mathcal{D}, |S| \ge n$ that is maximally resilient to the features uncovered by $\Phi$. In other words, 
for any identically-distributed train-test split of $S$, learning how to best exploit the features $\Phi$ on the training instances should not help models generalize to the held-out test set.
% Our approach allows for any choice of feature representation.

Let $\mathcal{M}$ denote a family of classification models (e.g., logistic regression, support vector machine classifier, or a particular neural architecture) that can be trained on subsets $S$ of $\mathcal{D} = (X,Y)$ using features $\Phi(X)$. 
We define the \emph{representation bias of $\Phi$ in $S$ w.r.t\ $\mathcal{M}$}, denoted $\mathcal{R}(\Phi, S, \mathcal{M})$, as the best possible out-of-sample classification accuracy achievable by models in $\mathcal{M}$ when predicting labels $Y$ using features $\Phi(X)$. 
Given a target minimum reduced dataset size $n$, the goal is to find a subset $S \subset \mathcal{D}$ of size at least $n$ that minimizes this representation bias in $S$ w.r.t.\ $\mathcal{M}$:
% \ashish{TODO: reconcile this $n$ with the stopping criterion in the actual algorithm, which uses a predictability / confidence threshold. doesn't look like there is a reliance on a pre-determined $n$, except for deciding how large $|T|$ should be when constructing $S$.}
\vspace*{2pt}
\begin{align}
    \label{eq:representation-bias}
    \argmin_{S \subset D,\, |S| \geq n} \mathcal{R}(\Phi, S, \mathcal{M})
\end{align}
Eq.~(\ref{eq:representation-bias}) corresponds to \emph{optimum bias reduction}, referred to as \afopt{}. We formulate $\mathcal{R}(\Phi, S, \mathcal{M})$ as the expected classification accuracy resulting from the following process. 
Let $q: 2^S \to [0,1]$ be a probability distribution over subsets $T = (X^T, Y^T)$ of $S$. 
The process is to randomly choose $T$ with probability $q(T)$, train a classifier $M_T \in \mathcal{M}$ on $S \setminus T$, and evaluate its classification accuracy $f_{M_{T}}\big(\Phi(X^T), Y^T\big)$ on $T$. 
The resulting accuracy on $T$ itself is a random variable, since the training set $S \setminus T$ is randomly sampled.
% and training the model $M_T$ on $D \setminus T$ may also involve randomization.  \ashish{dropping this, as we don't address it, e.g., via multiple training runs}
We define the expected value of this classification accuracy to be the representation bias:
\vspace*{2pt}
\begin{align}
    \label{eq:representation-bias-M}
    \mathcal{R}(\Phi, S, \mathcal{M})
    & \triangleq \E_{T \sim q} \left[f_{M_{T}}\big(\Phi(X^T), Y^T\big)\right]
\end{align}

% While this expression formalizes the intended objective function, it involves a large summation over subsets $T \subset S$ just to compute the representation bias present in a single set $S$. 
% It does not suggest a practical way to compute the minimization in Eq.~(\ref{eq:representation-bias}) without further considering each of the exponentially many subsets $S \subset D$ individually -- thus an optimal solution for Equation~(\ref{eq:representation-bias-M}) is intractable.
% To get around this difficulty, we reformulate the representation bias in $S$ as a sum factored over the $|S|$ individual instances $i \in S$. 
% This will allow us to efficiently decide whether or not to include $i$ in the targeted, reduced subset we are constructing.

The expectation in Eq.~(\ref{eq:representation-bias-M}), however, involves a summation over exponentially many choices of $T$ even to compute the representation bias for a \emph{single} $S$. This makes optimizing Eq.~(\ref{eq:representation-bias}), which involves a search over $S$, highly intractable. To circumvent this challenge, we refactor $\mathcal{R}(\Phi, S, \mathcal{M})$ as a sum over instances $i \in S$ of the \emph{aggregate} contribution of $i$ to the representation bias across all $T$. Importantly, this summation has only $|S|$ terms, allowing more efficient computation.
% The idea is to aggregate the contribution of each $i$ towards the representation bias expression across all random choices of the training set $D \setminus T$.
We call this the \emph{predictability score} $p(i)$ for $i$: on average, how reliably can label $y_i$ be predicted using features $\Phi(x_i)$ when a model from $\mathcal{M}$ is trained on a randomly chosen training set $S \setminus T$ not containing $i$.
% The higher the value of $p(i)$, the easier it is to correctly classify the instance $(x_i, y_i)$ using model family $\mathcal{M}$. This is the signal we will use to decide whether to include $i$ in the reduced subset $S$ we are constructing.
Instances with high predictability scores are undesirable as their feature representation can be exploited to confidently correctly predict such instances.
%\changedAshish{We would like all instances $i \in S$ to have low predictability scores.}

With some abuse of notation, for each $i \in S$, we denote $q(i) \triangleq \sum_{T \ni i} q(T)$ the marginal probability of choosing a subset $T$ that contains $i$. The ratio $\frac{q(T)}{q(i)}$ is then the probability of $T$ conditioned on it containing $i$. Let $f_{M_{T}}\big(\Phi(x_i), y_i\big)$ be the classification accuracy of $M_T$ on $i$. Then the expectation in Eq.~(\ref{eq:representation-bias-M}) can be written in terms of $p(i)$ as follows:
{\small
\begin{align*}
    % \E_{T \sim q} \left[ f_{M_{T}}(\Phi(X^T), Y^T) \right]
    & \sum_{T \subset S} q(T) \cdot \frac{1}{|T|} \sum_{i \in T} f_{M_{T}}\big(\Phi(x_i), y_i\big) \\
    & = \sum_{T \subset S} \, \sum_{i \in T} q(T) \cdot \frac{f_{M_{T}}\big(\Phi(x_i), y_i\big)}{|T|} \\
    & = \sum_{i \in S} \, \sum_{\substack{T \subset S\\T \ni i}} q(T) \cdot \frac{f_{M_{T}}\big(\Phi(x_i), y_i\big)}{|T|} \\
    & = \sum_{i \in S} q(i) \, \sum_{\substack{T \subset S\\T \ni i}} \frac{q(T)}{q(i)} \, \frac{f_{M_{T}}\big(\Phi(x_i), y_i\big)}{|T|} \\
    & = \sum_{i \in S} q(i) \, \E_{T \subset S,\, T \ni i} \left[ \frac{f_{M_{T}}\big(\Phi(x_i), y_i\big)}{|T|} \right] \\ 
    & = \ \ \sum_{i \in S} p(i)
\end{align*}
}
where
% , as before, $q(i)$ denotes the marginal probability of choosing a subset $T$ that contains $i$, and
$p(i)$ is the predictability score of $i$ defined as:
\begin{align}
    \label{eq:predictability-score}
    p(i) & \ \triangleq \ q(i) \cdot \E_{T \subset S,\, T \ni i} \left[ \frac{f_{M_{T}}\big(\Phi(x_i), y_i\big)}{|T|} \right]
\end{align}

While this refactoring works for any probability distribution $q$ with non-zero support on all instances, for simplicity of exposition, we assume $q$ to be the uniform distribution over all $T \subset S$ of a fixed size. This makes both $|T|$ and $q(i)$ fixed constants; in particular, $q(i) = \binom{|S|-1}{|T|-1} / \binom{|S|}{|T|} = \frac{|T|}{|S|}$.
% This reduces the predictability score expression of Eq.~(\ref{eq:predictability-score}) to the simplified variant $\tilde{p}(i)$:
This yields a simplified predictability score $\tilde{p}(i)$ and a factored reformulation of the representation bias from Eq.~(\ref{eq:representation-bias-M}):
\vspace*{1pt}
\begin{gather}
    \label{eq:predictability-score-approx}
    \tilde{p}(i) \triangleq \frac{1}{|S|} \, \E_{T \subset S,\, T \ni i} \left[ f_{M_{T}}\big(\Phi(x_i), y_i\big) \right] \\
% \end{align}
% Putting the pieces together, we have a factored reformulation of the representation bias in Eq.~(\ref{eq:representation-bias-M}):
% \begin{align}
    \label{eq:representation-bias-factored}
    \mathcal{R}(\Phi, S, \mathcal{M}) = \sum_{i \in S} \tilde{p}(i)
\end{gather}

Although this refactoring reduces the exponential summation underlying the expectation in Eq.~(\ref{eq:representation-bias-M}) to a linear sum, solving Eq.~(\ref{eq:representation-bias}) for optimum bias reduction (\afopt) remains challenging due to the exponentially many choices of $S$. However, the refactoring does enable computationally efficient heuristic approximations that start with $S = \mathcal{D}$ and iteratively filter out from $S$ the most predictable instances $i$, as identified by the (simplified) predictability scores $\tilde{p}(i)$ computed over the current candidate for $S$. 
%
%\papername{} adopts a \emph{greedy slicing approach}. Namely, it identifies the instances with the $k$ highest predictability scores, removes all of them from $S$, and repeats the process up to $\lfloor \frac{|\mathcal{D}|-n}{k} \rfloor$ times.
\papername{} adopts a greedy slicing approach.
Namely, it identifies the instances with the $k$ highest predictability scores, removes all of them from $S$, and repeats the process up to $\lfloor \frac{|\mathcal{D}|-n}{k} \rfloor$ times.
This can be viewed as a scalable and practical approximation of (intractable) \afopt{} for optimum bias reduction. In Appendix~\S\ref{sec:heuristics}, we compare three such heuristic approaches.
In all cases, we use a fixed training set size $|S \setminus T| = t < n$. Further, since a larger filtered set is generally desirable, we terminate the filtering process early (i.e., while $|S| > n$) if the predictability score for every $i$ falls below a pre-specified early stopping threshold $\tau \in [0,1]$.

\begin{algorithm}[t]
\small
\DontPrintSemicolon
\SetAlgoNoEnd
%\KwIn{twin sentences $\{T(t_1, t_2)_1^{N}\}$, threshold $h$}
\KwIn{dataset $D=(X, Y)$, pre-computed representation $\Phi(X)$, model family $\mathcal{M}$, target dataset size $n$, number of random partitions $m$, training set size $t < n$, slice size $k \leq n$, early-stopping threshold $\tau$}
%\KwOut{filtered twin sentences $\{T'(t_1, t_2)_1^{M}\}$}
\KwOut{reduced dataset $S$}
%$bias=True$\;
$S=D$\;

\While{$|S| > n$}{
    \tcp{Filtering phase}
    \ForAll{$i \in S$}{
        Initialize multiset of out-of-sample predictions $E(i) = \emptyset$\;
    }
    \For{iteration $j: 1..{m}$}{
        Randomly partition $S$ into $(T_j, S \setminus T_j)$ s.t.\ $|S \setminus T_j|=t$
        
        Train a classifier $\mathcal{L} \in \mathcal{M}$ on $\{(\Phi(x),y) \mid (x,y) \in S \setminus T_j\}$ ($\mathcal{L}$ is typically a linear classifier)
        
        % \ForAll{$i=({\bf x}, y) \in T_j$}{
        \ForAll{$i=(x, y) \in T_j$}{
            % Add $\mathcal{L}({\bf x})$ to $E(i)$\;
            Add the prediction $\mathcal{L}(\Phi(x))$ to $E(i)$\;
        }
    }
    %Compute $score$ for all elements of $D'$ as the ratio of correct predictions\\
    % \ForAll{$i=({\bf x}, y) \in S$}{
    \ForAll{$i=(x, y) \in S$}{
        Compute the predictability score $\tilde{p}(i) = |\{\hat{y} \in E(i)~\textit{s.t.}~\hat{y} = y\}|\, /\, |E(i)|$\;
    }
    Select up to $k$ instances $S'$ in $S$ with the highest predictability scores subject to $\tilde{p}(i) \geq \tau$
    
    $S = S \setminus S'$\;
    
    \If{$|S'| < k$}{
        %$bias=False$\;
        \textbf{break}
    }
}
\Return $S$
\caption{\papername{}}
\label{alg:filter}
\end{algorithm}
% UNCOMMENT TO MAKE FIG BIGGER

\begin{figure*}[ht]
    \centering
    \includegraphics[width=0.83\linewidth]{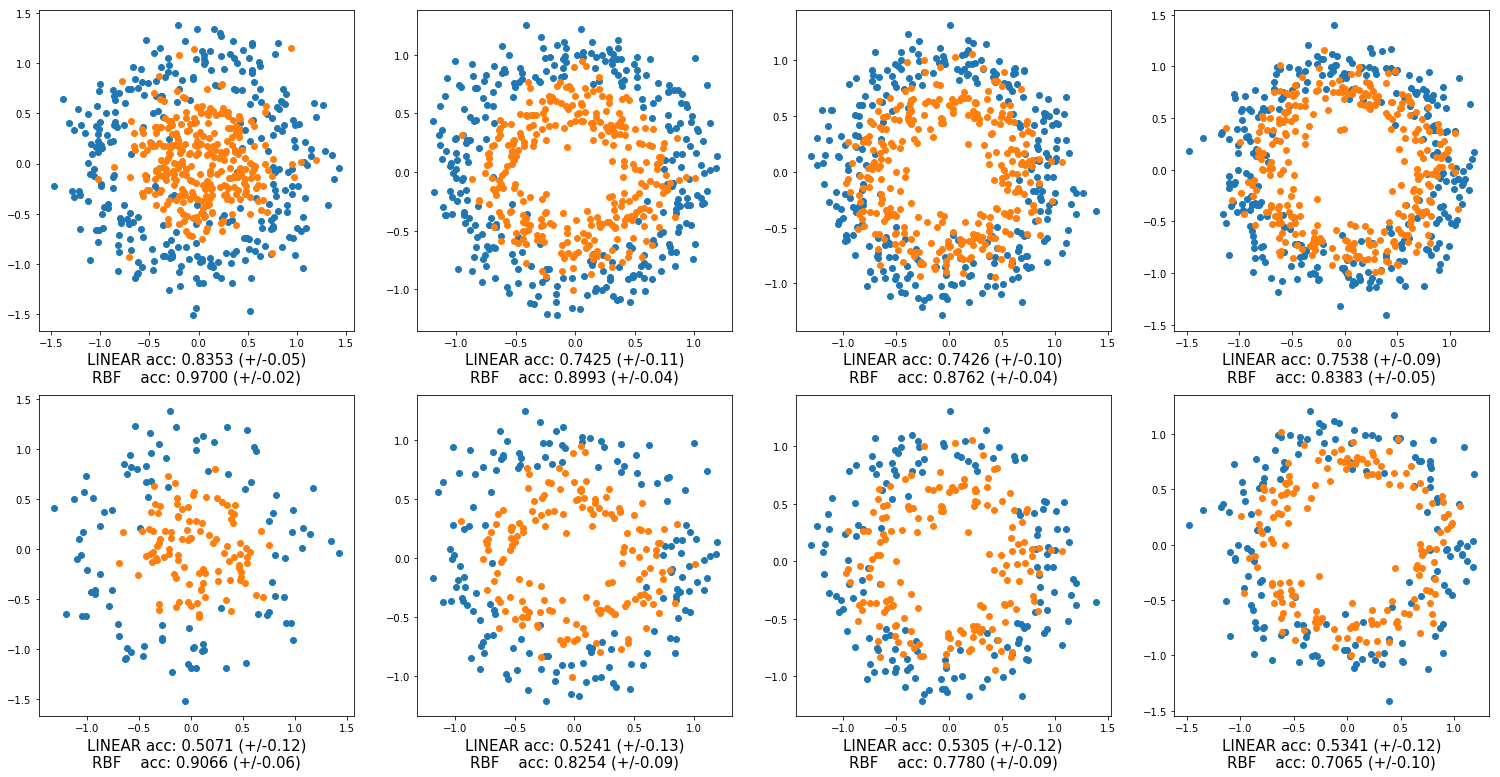}
    \caption{Four sample biased datasets as input to \papername (top). Blue and orange indicate two different classes. Only the original two dimensions are shown, not the bias features. For the leftmost dataset with the highest separation, we flip some labels at random, so even an RBF kernel cannot achieve perfect performance. \papername{} makes the data more challenging for the models (bottom).} 
     \label{fig:synthetic}
\end{figure*}

% COMMENT TO MAKE FIG BIGGER
% \begin{figure}[ht]
%     \centering
%     \includegraphics[scale=0.3]{sections/figures/synthetic-2.png}
%     \caption{Two sample biased datasets as input to \papername (top). Blue and orange indicate two different classes. Only the original two dimensions are shown, not the bias features. For the dataset on the left, with the highest separation, we flip some labels at random, so even an RBF kernel cannot achieve perfect performance. \papername{} makes the data more challenging for the models (bottom). Also see Appendix \S\ref{sec:syn_results} for more details.} 
%      \label{fig:synthetic}
% \end{figure}

\textbf{Implementation} Algorithm~\ref{alg:filter} provides an implementation of \papername{}.
The algorithm takes as input a dataset $D = (X,Y)$, a representation $\Phi(X)$ we are interested in minimizing the bias in, a model family $\mathcal{M}$ (e.g., linear classifiers), a target dataset size $n$, size $m$ of the support of the expectation in Eq.~(\ref{eq:predictability-score-approx}), training set size $t$ for the classifiers, size $k$ of each slice, and an early-stopping filtering threshold $\tau$.
Importantly, for efficiency, $\Phi(X)$ is provided to \papername{} in the form of \emph{pre-computed} embeddings for all of $X$.
%\swabha{NEW NEW NEW}
%To obtain $\Phi(X)$, we train a first model, using a small fraction of the data, such that the performance of this model is within a reasonable limit of the full model performance.
%We used preliminary experiments to determine the size of this small fraction of samples, and do not reuse these for the rest of our experiments.
%We determine the size of this small fraction of samples based on the learning curve in low-data regime, and do not reuse these for the rest of our experiments.
In practice, to obtain $\Phi(X)$, we train a first ``warm-up'' model on a small fraction of the data based on the learning curve in low-data regime, and do not reuse this data for the rest of our experiments.
Moreover, this fraction corresponds to the training size $t$ for \papername{} and it remains unchanged across iterations.
%Based on the performance and learning curve of the initial model, we also set the filtering threshold $\tau$.
We follow the iterative filtering approach, starting with $S = D$ and iteratively removing some instances with the highest predictability scores using the greedy slicing strategy.
Slice size $k$ and number of partitions $m$ are determined by the available computation budget.
%, since we want to run \papername{} for only a reasonable number of iterations, on a machine with a fixed number of CPU cores. 

At each filtering phase, we train models (linear classifiers) on $m$ different random partitions of the data, and collect their predictions on their corresponding test set. For each instance $i$, we compute its \emph{predictability score} as the ratio of the number of times its label $y_i$ is predicted correctly, over the total number of predictions for it. 
We rank the instances according to their predictability score and use the greedy slicing strategy of removing the top-$k$ instances whose score is not less than the early-stopping threshold $\tau$. 
We repeat this process until fewer than $k$ instances pass the $\tau$ threshold in a filtering phase or fewer than $n$ instances remain.
Appendix~\S\ref{sec:aflite_hyperparams} provides details of hyperparameters used across different experimental settings, to be discussed in the following sections.

\section{Synthetic Data Experiments}
\label{sec:exp_syn}

% \begin{table*}
% \small
% \centering
% \begin{tabular}{lcccccccccc}
% \toprule
%  & \multicolumn{4}{c}{HANS} & \multicolumn{3}{c}{NLI-Diagnostics} & \multicolumn{3}{c}{Adversarial-NLI} \\
% \cmidrule(lr){2-5}\cmidrule(lr){6-8}\cmidrule(lr){9-11}
% Model & \emph{All} & \emph{Lex.} & \emph{Subseq.} & \emph{Constit.} & \emph{All} & \emph{Logic} & \emph{Knowl.} & \emph{Rd1} & \emph{Rd2} & \emph{Rd3} \\ \midrule
% $\text{RoBERTa}$    & 70.7 & 84.4 & 35.4 & 13.4 & 59.3 & 52.8 & 48.9 & 58.5 & 48.3 & 50.1  \\
% $\text{RoBERTa-AFlite}$     & \textbf{74.5} & \textbf{96.3} & \textbf{56.6} & \textbf{57.4} & \textbf{62.0} & \textbf{53.2} & \textbf{57.7} & \textbf{65.1} & \textbf{49.1} & \textbf{52.8} \\
% \bottomrule
% \end{tabular}
% \caption{
% SNLI accuracy (\%) on three out-of-distribution evaluation tasks, comparing RoBERTa-large models pre-trained on the original SNLI data, and on \papername{}-filtered data. On the HANS dataset, both models are evaluated on \emph{All}, as well as on the non-entailment cases of the three syntactic heuristics (\emph{Lexical overlap}, \emph{Subsequence}, and \emph{Constituent}). The NLI-Diagnostics dataset is broken down into the full dataset (\emph{All}), as well as the instances requiring logical reasoning (\emph{Logic}) and the ones requiring world and commonsense knowledge (\emph{Knowledge}). For Adversarial NLI, we finetuned both models on the in-distribution training data for each round (\emph{Rd1}, \emph{Rd2}, and \emph{Rd3}).
% }
% \label{tab:snli_gen}
% \end{table*}

\begin{table*}
\small
\centering
\caption{
Zero-shot SNLI accuracy on three out-of-distribution evaluation tasks, comparing \roberta-large models trained on the original SNLI data ($D$, size 550k), \papername{}-filtered data (\robfilt, size 182k), and on a random subset with the same size as the filtered data ($D_{182k}$).
The reported accuracy is averaged across 5 random seeds, and the subscript denotes standard deviation.
On the HANS dataset, all models are evaluated on the non-entailment cases of the three syntactic heuristics (\emph{Lexical overlap}, \emph{Subsequence}, and \emph{Constituent}). 
The NLI-Diagnostics dataset is broken down into the instances requiring world and commonsense knowledge (\emph{Knowl.}), logical reasoning (\emph{Logic}), predicate-argument structures (\emph{PAS}), or lexical semantics (\emph{LxS.}).  
Stress tests for NLI are further categorized into \emph{Competence}, \emph{Distraction} and \emph{Noise} tests.
}
\begin{tabular}[t]{lcccccccccc}
\toprule
 & \multicolumn{3}{c}{HANS} 
 & \multicolumn{4}{c}{NLI-Diagnostics} 
 & \multicolumn{3}{c}{Stress}\\
\cmidrule(lr){2-4}\cmidrule(lr){5-8}\cmidrule(lr){9-11}\\
& \emph{Lex.} & \emph{Subseq.} & \emph{Constit.} 
& \emph{Knowl.} & \emph{Logic} & \emph{PAS} & {LxS.}
& \emph{Comp.} & \emph{Distr.} & \emph{Noise} \\ \midrule
$D$
& 88.4$_{2.2}$ & 28.2$_{3.4}$ & 21.7$_{7.1}$ 
& 51.8$_{1.6}$ & 57.8$_{1.7}$ & \bf{72.6$_{1.3}$} & 65.7$_{1.9}$
& 77.9$_{2.5}$ & \bf{73.5$_{2.9}$} & \bf{79.8$_{0.8}$} \\
$D_{182k}$
& 56.6$_{14.7}$ & 19.6$_{5.6}$ & 13.8$_{2.9}$ 
& \bf{56.4}$_{0.8}$ & 53.9$_{1.5}$ & 71.2$_{1.1}$ & 65.6$_{1.7}$
& 68.4$_{3.0}$ & 73.0$_{3.0}$ & 78.6$_{0.4}$ \\
\robfilt 
%& 87.8$_{3.6}$ & 36.5$_{8.7}$ & 27.9$_{12.3}$ 
& \bf{94.1}$_{3.5}$ & \bf{46.3}$_{6.0}$ & \bf{38.5}$_{15.2}$ 
& 53.9$_{1.6}$ & \bf{58.7$_{1.2}$} &69.9$_{0.9}$ & \bf{66.5$_{1.7}$}
& \bf{79.1$_{1.0}$} & 72.0$_{1.8}$ & 79.5$_{0.4}$  \\
% & 54.0$_{1.2}$ & 56.9$_{1.5}$ & 71.4$_{0.3}$ & 66.4$_{2.3}$
% & 75.3$_{4.7}$ & 71.5$_{1.8}$ & 79.0$_{0.6}$  \\
\bottomrule
\end{tabular}
\label{tab:snli_gen}
\end{table*}

\begin{table}
\small
\centering
\caption{
SNLI accuracy on Adversarial NLI using \roberta-large models pre-trained on the original SNLI data ($D$, size 550k) and on \papername{}-filtered data (\robfilt, size 182k).
Both models were finetuned on the in-distribution training data for each round (\emph{Rd1}, \emph{Rd2}, and \emph{Rd3}).
}
\begin{tabular}[t]{lccc}
\toprule
 & \multicolumn{3}{c}{Adversarial-NLI}\\
\cmidrule(lr){2-4}
& \emph{Rd1} & \emph{Rd2} & \emph{Rd3} \\
$D$
& 58.5 & 48.3 & 50.1 \\
\robfilt
&\bf 65.1 & \bf 49.1 & \bf 52.8 \\
% $D_{182k}$
% & \todo{} & \todo{} & \todo{} \\
\bottomrule
\end{tabular}
\label{tab:snli_adversarial}
\end{table}

We present experiments under a synthetic setting, to evaluate whether \papername{} successfully removes examples with spurious correlations from a dataset.
We synthesize a dataset comprising two-dimensional data, arranged in concentric circles, at four different levels of separation, as shown in Figure~\ref{fig:synthetic}.
The label (color) indicates the circular region the data point is situated in.
As is evident, a linear function is inadequate for separating the two classes; it requires a more complex non-linear model such as a support vector machine (SVM) with a radial basis function (RBF) kernel. 

To simulate spurious correlations in the data, we add class-specific artificially constructed features (biases) sampled from two different Gaussian distributions.
These features are only added to $75\%$ of the data in each class, while for the rest of the data, we insert random (noise) features.
The bias features make the task solvable through a linear function.
Furthermore, for the first dataset, with the largest separation, we flipped the labels of some biased samples, making the data slightly adversarial even to the RBF.
Both models can clearly leverage the biases, and demonstrate improved performance over a baseline without biases.\footnote{We use standard implementations from scikit-learn:~\url{https://scikit-learn.org/stable/}.}

Once we apply \papername{}, as expected, the number of biased samples is reduced considerably, making the task hard once again for the linear model, but still solvable for the non-linear one. 
The filtered dataset is shown in the bottom half of Fig. \ref{fig:synthetic}, and the captions indicate the performance of a linear and an SVM model (detailed results for each are provided in Appendix \S\ref{sec:syn_results}, for better visibility).
Under each separation level, our results show that \papername{} indeed 
% lowers the performance of models relying on biases by removing 
removes examples with spurious correlations from a dataset.
Moreover, \papername removes most of the flipped examples in the first dataset.

\begin{table*}
\small
\centering
\caption{\emph{Dev} accuracy ($\%$) on the original SNLI dataset $D$ and the datasets obtained through different \papername-filtering and other baselines.
$D_{92k}$ indicates a randomly subsampled train dataset of the same size as \robfilt.
$\Delta$ indicates the difference in performance (or size, last row) between the full model and the model trained on \robfilt.
}
\begin{tabular}[t]{lcccccc}
\toprule
&  \multicolumn{5}{c}{Train Data} & \\
                            \cmidrule{2-6} 
%              & \multicolumn{5}{c}{SNLI Dataset} \\
%\cmidrule(lr){2-6}
% Model            & Full   & Random & ESIM+GLoVe-\papername & BERT-\papername & RoBERTa-AFLite\\
Model            & $D$   & $D_{92k}$ & $D(\Phi_{\textit{ESIM+GLoVe}})$ & \bertfilt & \robfilt & $\Delta$ \\
\midrule
\textit{ESIM+ELMo} {\small \citep{Peters2018DeepCW}} & 88.7 & 86.0 & 61.5 & 54.2 & 51.9 & -36.8\\
\bert{\small~\citep{Devlin2019Bert}} & 91.3 & 87.6 & 74.7 & 61.8 & 57.0 & -34.3 \\
\roberta {\small~\citep{Liu2019RoBERTaAR}} & 92.6 & 88.3 & 78.9 & 71.4 & 62.6 & -30.0\\
\midrule
Max-PPMI & 54.5 & 52.0 & 41.1 & 41.5 & 41.9 & -12.6\\
\bert\hyp & 71.5 & 70.1 & 52.3 & 46.4 & 48.4 & -23.1\\
\roberta\hyp  & 72.0 & 70.4 & 53.6 & 49.5 & 48.5 & -23.5\\
\midrule
\emph{Human performance} & 88.1 & 88.1 & 82.3 & 80.3 & 77.8 & -10.3\\
\emph{Training set size} & 550k & 92k & 138k & 109k & 92k & -458k\\
\bottomrule
\end{tabular}
%\mpcomment{Do we have human performance for the datasets? If so, add them as an additional row in the table.  Maybe also a row with number of training instances in each dataset.}
%\mpcomment{Both this table and the MNLI/QNLI table would also benefit from a random baseline row if it changes significantly from the original data.}
%\mpcomment{Also need to add a column for the random subsample baseline (performance with a random subsample of SNLI that is the same size as the AF'ed dataset, to control for size}
\label{tab:SNLIresults}
\end{table*}
\section{NLP Experiments}
\label{sec:exp_nlp}

As our first real-world data evaluation for \papername, we consider out-of-domain and in-domain generalization for a variety of language datasets.
The primary task we consider is natural language inference (NLI) on the Stanford NLI dataset \citep[SNLI]{Bowman2015ALA}.
Each instance in the NLI task consists of a premise-hypothesis sentence pair, the task involves predicting whether the hypothesis either \textit{entails}, \textit{contradicts} or is \textit{neutral} to the premise.

\textbf{Experimental Setup}
%\mpcomment{Someone add details of feature pretraining}.
%\mpcomment{also add details about target dataset size and class balance in the filtered data, perhaps in appendix}
We use feature representations from \roberta-large, \robphi \citep{Liu2019RoBERTaAR}, a large-scale pretrained masked language model. 
This is extracted from the final layer before the output layer, trained on a random $10\%$ sample (warm-up) of the original training set.
% These features are pre-computed for all remaining instances and the instances used for the warm-up training above are discarded for the rest of the process. 
The resultant filtered NLI dataset, \robfilt, is compared to the original dataset $D$ as well as a randomly subsampled dataset $D_{182k}$, with the same sample size as \robfilt, amounting to only a third of the full data $D$.
The same \roberta-large architecture is used to train the three NLI models.

% \papername assigns a predictability score to all samples in a dataset, resulting in an ordering of the data.
% Filtering out examples from the head of the data distribution based on this order yields more accurate benchmarks for measuring true model performance.
% On the other hand, transferability to out-of-distribution data would involve a greater balance between examples from the head and the tail ends of the data distribution.
% Hence we evaluate \papername{} for generalization using a larger filtered subset, amounting to only a third of the full training data.

\subsection{Out-of-distribution Generalization}
\label{sec:snli-gen}

As motivated in Section \S\ref{sec:intro}, large-scale architectures often learn to solve datasets rather than the underlying task by overfitting on unintended correlations between input and output in the data.
However, this reliance might be hurtful for generalization to out-of-distribution examples, since they may not contain the \textit{same} biases.
% If a model is trained on data free of spurious biases, we hypothesize that it would be able to retain its generalization performance.
We evaluate \papername{} for this criterion on the NLI task.

\citet{gururangan-etal-2018-annotation}, among others, showed the existence of certain annotation artifacts (lexical associations etc.) in SNLI which make the task considerably easier for most current methods.
This spurred the development of several out-of-distribution test sets which carefully control for the presence of said artifacts.
% in order to truly estimate the capability of NLI models in the wild.
We evaluate on four such out-of-distribution datasets: HANS \cite{mccoy-etal-2019-right}, NLI Diagnostics \cite{Wang2018GLUEAM}, Stress tests \cite{naik-etal-2018-stress} and Adversarial NLI \cite{Nie2019-ri} (c.f. Appendix \S\ref{sec:nli-ood} for details).
% SNLI stress tests \citep{naik-etal-2018-stress} contain rule-based constructions, to account for robustness to biases, via replacement of words and entities in the hypothesis and premise.
Given that these benchmarks are collected independently of the original SNLI task, the biases from SNLI are less likely to carry over.\footnote{However, these benchmarks might contain their own biases \citep{liu-etal-2019-inoculation}.}

Table \ref{tab:snli_gen} shows results on three out of four diagnostic datasets (HANS, NLI-Diagnostics and Stress), where we perform a zero-shot evaluation of the models. 
Models trained on SNLI-\papername{} consistently exceed or match the performance of the full model on the benchmarks above, up to standard deviation.
% thus improving the generalization ability due to reduced spurious biases.
% The slight decrease in performance in some cases can be attributed to the reduced amount of training data.
To control for the size, we compare to a baseline trained on a random subsample of the same size ($D_{182k}$).
\papername{} models report higher generalization performance suggesting that the filtered samples are more informative than a random subset.
In particular, \papername substantially outperforms challenging examples on the HANS benchmark, which targets models purely relying on lexical and syntactic cues.  
% Similarly, our model performs better on the instances in NLI-Diagnostics that require logical reasoning and commonsense knowledge, as opposed to instances that can be solved through lexical entailment alone. }
Table \ref{tab:snli_adversarial} shows results on the Adversarial NLI benchmark, which allows for evaluation of transfer capabilities, by finetuning models on each of the three training datasets (\emph{Rd1}, \emph{Rd2} and \emph{Rd3}).
A \roberta-large model trained on SNLI-\papername surpasses the performance in all three settings.

%\todo{More insights on each model. Question: do we know to go into the details of the performance on the lexical semantics part of NLI-Diagnostics?}

\subsection{In-distribution Benchmark Re-estimation}
\label{sec:snli-id}

\papername{} additionally provides a more accurate estimation of the benchmark performance on several tasks.
Here we simply lower the \papername early-stopping threshold, $\tau$ in order to filter most biased examples from the data, resulting in a stricter benchmark with 92k train samples.

\paragraph{SNLI}
In addition to \roberta-large, we consider here pre-computed embeddings from \bert-large \citep{Devlin2019Bert}, and \glove \citep{Pennington2014GloveGV},  resulting in three different feature representations for SNLI: \bertphi, \robphi from \roberta-large \citep{Liu2019RoBERTaAR}, and $\Phi_{\textit{ESIM+GLoVe}}$ which uses the ESIM model \citep{Chen2016EnhancedLF} with \glove embeddings.
Table \ref{tab:SNLIresults} shows the results for SNLI. 
In all cases, applying \papername{} substantially reduces overall model accuracy, with typical drops of 15-35\% depending on the models used for learning the feature representations and those used for evaluation of the filtered dataset. 
In general, performance is lowest when using the strongest model (\roberta) for learning feature representations.
Results also highlight the ability of weaker adversaries to produce datasets that are still challenging for much stronger models with a drop of 13.7\% for \roberta using $\Phi_{\textit{ESIM+GLoVe}}$ as feature representation. 

% It might seem unsurprising that reducing the size of the training set results in lower performance. 
To control for the reduction in dataset size by filtering, we randomly subsample $D$, creating $D_{92k}$ whose size is approximately equal to that of \robfilt. 
All models achieve nearly the same performance as their performance on the full dataset -- even when trained on just one-fifth the original data. 
This result further highlights that current benchmark datasets contain significant redundancy within its instances.

We also include two other baselines, which target known dataset artifacts in NLI.
The first baseline uses Point-wise Mutual Information (PMI) between words in a given instance and the target label as its only feature. 
Hence it captures the extent to which datasets exhibit word-association biases, one particular class of spurious correlations. 
While this baseline is relatively weaker than other models, its performance still reduces by nearly 13\%  on the \robfilt dataset.
The second baseline trains only the hypothesis of an NLI instance ({\hyp{}}).
Such partial input baselines \cite{gururangan-etal-2018-annotation} capture reliance on lexical cues only in the hypothesis, instead of learning a semantic relationship between the hypothesis and premise.
This reduces performance by almost 24\% before and after filtering with \roberta.
\papername, which is agnostic to any particular known bias in the data, results in a drop of about 30\% on the same dataset, indicating that it might be capturing a larger class of spurious biases than either of the above baselines.

Finally, to demonstrate the value of the iterative, ensemble-based \papername{} algorithm, we compare with a baseline where using a single model, we filter out the most predictable examples in a single iteration --- a non-iterative, single-model version of \papername{}. 
%This baseline uses \roberta features for assigning predictability scores to all samples. 
A \roberta-large model trained on this subset (of the same size as \robfilt) achieves a dev accuracy of $72.1\%$. 
Compared to the performance of \roberta on \robfilt ($62.6\%$, see Table~\ref{tab:SNLIresults}), it makes this baseline a sensible yet less effective approach.
In particular, this illustrates the need for an iterative procedure involving models trained on multiple partitions of the remaining data in each iteration.

\vspace{-2pt}

\begin{table}
\small
\centering
\caption{\emph{Dev} accuracy ($\%$) on the original ($D$) and \papername-filtered (\robfilt) MultiNLI-matched and QNLI datasets. 
The {\partialinput} baselines show models trained on only \emph{Hypotheses} for MultiNLI instances and only \emph{Answers} for QNLI. $\Delta$ indicates the difference in accuracy of the full model and the filtered model.
}
%\begin{tabular}[t]{lccc}
%\toprule
%               & \multicolumn{2}{c}{MNLI} \\
%\cmidrule(lr){2-3}
%Model            & $D$    & \robfilt & $\Delta$ \\
%\midrule
%\bert & 86.6 & 55.8 & 30.8\\
%\roberta  & 90.3 & 66.2 & 24.1\\
%\midrule
%bert\partialinput  & 59.7 & 43.2 &  16.5\\
%\roberta\partialinput & 60.3 & 44.4 & 15.9 \\
%\midrule
%& \multicolumn{2}{c}{QNLI} \\
%\cmidrule(lr){2-3}
%Model          & $D$    & \robfilt & $\Delta$ \\
%          \midrule
%\bert &  92.0 & 63.5 & 28.5 \\
%\roberta  & 93.7 & 77.7 & 16.0\\
%\midrule
%\bert\partialinput  & 62.6 & 56.6 & 6.0\\
%\roberta\partialinput & 63.9 & 59.4 & 4.5\\
%\bottomrule
%\end{tabular}
\begin{tabular}[t]{llccc}
\toprule
        &       & \multicolumn{2}{c}{Train Data} \\
\cmidrule(lr){3-4}
Task & Model            & $D$    & \robfilt & $\Delta$ \\
\midrule
\multirow{4}{*}{QNLI} & \bert & 86.6 & 55.8 & -30.8\\
& \roberta  & 90.3 & 66.2 & -24.1\\
\cmidrule(lr){2-5}
& \bert{}\partialinput  & 59.7 & 43.2 &  -16.5\\
& \roberta{}\partialinput & 60.3 & 44.4 & -15.9 \\
\midrule
\multirow{4}{*}{\parbox{0.6cm}{Multi-NLI}} &\bert &  92.0 & 63.5 & -28.5 \\
&\roberta  & 93.7 & 77.7 & -16.0\\
\cmidrule(lr){2-5}
&\bert{}\partialinput  & 62.6 & 56.6 & -6.0\\
&\roberta{}\partialinput & 63.9 & 59.4 & -4.5\\
\bottomrule
\end{tabular}
\label{tab:MNLI_QNLIresults}
\end{table}

\vspace{-2pt}
\paragraph{MultiNLI and QNLI}
We evaluate the performance of another large-scale NLI dataset multi-genre NLI \citep[MultiNLI]{williams-etal-2018-broad}, and the QNLI dataset \cite{Wang2018GLUEAM} which is a sentence-pair classification version of the SQuAD \citep{Rajpurkar2016SQuAD10} question answering task.\footnote{QNLI is stylized as an NLI classification task, where the task is to determine whether or not a sentence contains the answer to a question.}
Results before and after \papername are reported in Table \ref{tab:MNLI_QNLIresults}. Since \roberta resulted in the largest drops in performance across the board in SNLI, we only experiment with \roberta as adversary for MultiNLI and QNLI.  
While \roberta achieves over $90\%$ on both original datasets, its performance drops to $66.2\%$ for MultiNLI and to $77.7\%$ for QNLI on the filtered datasets. 
Similarly, partial input baseline performance also decreases substantially on both dataset compared to their performance on the original dataset. 
Overall, our experiments indicate that \papername{} consistently results in reduced accuracy on the filtered datasets across multiple language benchmark datasets, even after controlling for the size of the training set.

Table~\ref{tab:SNLIresults} shows that human performance on SNLI-\papername is lower than that on the full SNLI.\footnote{Measured based on five annotator labels provided in the original SNLI validation data.}
This indicates that the filtered dataset is somewhat harder even for humans, though to a much lesser degree than any model.
Indeed, removal of examples with spurious correlations could inadvertently lead to removal of genuinely easy examples; this might be a limitation of a model-based bias reduction approach such as \papername (see Appendix \S\ref{sec:qualitative_nli} for a qualitative analysis).
Future directions for bias reduction techniques might involve additionally accounting for unaltered human performance before and after dataset reduction.
\section{Vision Experiments}
\label{sec:exp_vision}

We evaluate \papername{} on image classification through ImageNet (ILSVRC2012) classification. 
On ImageNet, we use the state-of-the-art EfficientNet-B7 model \citep{tan2019efficientnet} as our core feature extractor $\Phi_\text{EN-B7}$. 
The EfficientNet model is learned from scratch on a fixed 20\% sample of the ImageNet training set, using RandAugment data augmentation \citep{cubuk2019randaugment}. 
We then use the 2560-dimensional features extracted by EfficientNet-B7 as the underlying representation for \papername{} to use to filter the remaining dataset, and stop when data size is 40\% of ImageNet.

\paragraph{Adversarial Image Classification}

In Table~\ref{tab:imagenet-gen}, we report performance of image classification models on ImageNet-A, a dataset with out-of-distribution images \cite{hendrycks2019natural}. 
As shown, all EfficientNet models struggle on this task, even when trained on the entire ImageNet.\footnote{Notably, there is a large difference in the degree of out-of-distribution generalization performance for NLP and vision tasks. NLP tasks benefit from the availability of pretrained representations from large language models, such as \roberta. In vision, however, while (pre)training on ImageNet alone is often sufficient to learn competitive features, such strong pretrained representations are not available. Moreover, ImageNet has many classes and a skewed distribution of data \cite{vodrahalli2018all}. Hence, it is considerably harder to find a smaller subset of data which generalizes well to adversarial challenge sets, such as ImageNet-A.}
However, we find that training on \papername{}-filtered data leads to models with greater generalization, in comparison to training on a randomly sampled ImageNet of the same size, leading to up to 2\% improvement in performance.

% In particular, we find that training on the 40\% hardest ImageNet-train examples (as judged by \papername{}) leads to 2\% greater performance versus training on 40\% random examples.

\begin{table}
\small
\centering
\caption{Top-1 accuracy on ImageNet-A \cite{hendrycks2019natural}, an adversarial evaluation set for image classification. The most powerful model EfficientNet-B7 improves by 2\% on out-of-distribution ImageNet-A images when trained on \papername{}-filtered data $D(\Phi_\text{EN-B7})$.}
\begin{tabular}[t]{lcc}
\toprule
& \multicolumn{2}{c}{Model}\\
\cmidrule{2-3}
Train Data                   &  EfficientNet-B5 & EfficientNet-B7  \\ 
\midrule
$D$                     & 16.5   & 20.6 \\
$D_{40\%}$              & 5.9  & 8.5    \\
$D(\Phi_\text{EN-B7})$  & 7.2 & 10.4    \\
\bottomrule
\end{tabular}
\label{tab:imagenet-gen}
\end{table}

\paragraph{In-distribution Image Classification}

In Table~\ref{tab:imagenet}, we present ImageNet accuracy across the EfficientNet and ResNet \citep{he2016deep} model families before and after filtering with \papername{}.
%  When lowering the size of the training set -- down to 20\% of the original, we find a large drop in performance. 
%  The EfficientNet models seem to suffer less -- from $84\%$ to $73\%$ on EfficientNet-B7 versus $80.6\%$ to $56.5\%$ on ResNet-152. 
For evaluation, the Imagenet-\papername{} filtered validation set is much harder than the standard validation set (also see Figure~\ref{fig:imagenetexamples}).
While the top performer after filtering is still EfficientNet-B7, its top-1 accuracy drops from 84.4\% to 63.5\%.
A model trained on a randomly filtered subsample of the same size though suffers much less, most likely due to reduction in training data. 
% The large drop persists even when training the model on Imagenet-\papername{} filtered data. %This is despite controlling for dataset size, as well as discrepancy between the training and validation sets.

Overall, these results suggest that image classification -- even within a subset of the closed world of ImageNet -- is far from solved. 
These results echo other findings that suggest that common biases that naturally occur in web-scale image data, such as towards canonical poses \citep{alcorn2019strike} or towards texture rather than shape \citep{geirhos2018imagenet}, are problems for ImageNet-trained classifiers. 

\begin{table}
\small
\centering
\caption{Results on ImageNet, in Top-1 accuracy (\%). 
We consider training on the 40\% challenging instances, as filtered by \papername{} ($D(\Phi_\text{EN-B7})$), and compare this to a random 40\% subsample of ImageNet ($D_{40\%}$). 
We report results on the ImageNet validation set before and after filtering with \papername{}.  
$\Delta$ indicates the difference in accuracy when trained on the full data and the filtered data.
Notably, evaluating on \papername{}-filtered ImageNet is much harder---resulting in a drop of nearly 21 percentage points in accuracy for the strongest model. 
}
\begin{tabular}[t]{lcccccc}
\toprule
                            &  \multicolumn{3}{c}{Train Data} & \\
                            \cmidrule{2-4} 
Model                       &  $D$  & $D_{40\%}$ & $D(\Phi_\text{EN-B7})$ & $\Delta$\\ \midrule
{EfficientNet-B0}           & 76.3     & 69.6        & 50.2 & -26.1 \\
{EfficientNet-B3}           & 81.7     & 75.1        & 57.3 & -24.4 \\
{EfficientNet-B5}           & 83.7     & 78.6        & 62.2 & -21.5\\
{EfficientNet-B7}           & 84.4     & 78.8        & 63.5 & -20.9\\
\midrule
{ResNet-34}                 &  78.4    & 65.9        & 46.9 & -31.5\\
{ResNet-50}                 &  79.2    & 68.9        & 50.1 & -29.1\\
{ResNet-101}                &  80.1    & 70.1        & 52.2 & -27.9\\
{ResNet-152}                &  80.6    & 71.0        & 53.3 & -27.3\\
%  & \multicolumn{2}{c}{100\% Train, Original Val} & \multicolumn{2}{c}{20\% Train, Original Val} & \multicolumn{2}{c}{\papername{}} \\
% \cmidrule(lr){2-3}\cmidrule(lr){4-5}\cmidrule(lr){6-7}
% Model & Top-1 & Top-5 & Top-1 & Top-5 & Top-1 & Top-5 \\ \midrule
% EfficientNet-B0 & 76.3 & 93.2 & 58.5 & 81.2 & 18.1 & 48.1 \\
% EfficientNet-B2 & 79.8 & 94.9 &  60.9 & 82.8 & 20.7 & 53.1 \\
% EfficientNet-B4 & 82.6 & 96.3 &  64.4 & 85.8 & 23.3 & 58.8 \\
% EfficientNet-B7 & \textbf{84.4} & \textbf{97.1} & \textbf{73.8} & \textbf{90.8} & \textbf{24.5} & \textbf{60.6}  \\ \midrule
% ResNet-34 & 78.4 & 94.4 & 51.8 & 74.3 & 11.1 & 30.2 \\
% ResNet-50 & 79.2 & 94.7 & 53.2 & 75.5 & 12.2 & 30.2 \\
% ResNet-101 & 80.1 & 95.4 & 55.6 & 77.5 & 12.3 & 32.1 \\
% ResNet-152 & 80.6 & 95.5 & 56.5 & 78.2 & 13.2 & 33.8 \\
\bottomrule
\end{tabular}
\label{tab:imagenet}
\end{table}

\section{Related Work}
\label{sec:related}

\paragraph{Adversarial Filtering}
\papername{} is related to \citet{Zellers2018SWAGAL}'s adversarial filtering (AF) algorithm, yet distinct in two key ways: it is (i) much more broadly applicable (by not requiring over generation of data instances), and (ii) considerably more lightweight (by not requiring re-training a model at each iteration of AF). 
Variants of this AF approach have recently been used to create other datasets such as HellaSwag \citep{Zellers2019HellaSwagCA} and Abductive NLI \citep{Bhagavatula2019AbductiveCR} by iteratively perturbing dataset instances until a target model cannot fit the resulting dataset. 
While effective, these approaches run into three main pitfalls. 
First, dataset curators need to explicitly devise a strategy of collecting or generating perturbations of a given instance. 
Second, the approach runs the risk of distributional bias where a discriminator can learn to distinguish between machine generated instances and human-generated ones. 
Finally it requires re-training a model at each iteration, which is computationally expensive especially when using a large model such as \bert as the adversary. 
In contrast, \papername{} focuses on addressing dataset biases from existing datasets instead of adversarially perturbing instances. 
\papername{} was earlier proposed by \citet{sakaguchi2019winogrande} to create the Winogrande dataset. 
This paper presents more thorough experiments, theoretical justification and results from generalizing the proposed approach to multiple popular NLP and Vision datasets.

\paragraph{Data Selection for Debiased Representations}
\citet{Li2019REPAIRRR} recently proposed REPAIR, a method to remove representation bias by dataset resampling. 
% While resampling is a common technique for balancing datasets, t
The motivation in REPAIR is to learn a probability distribution over the dataset that favors instances that are hard for a given representation. 
% This approach targets how to train better, less-biased models as opposed to creating datasets with fewer artifacts. 
In contrast to \papername{}, the implementation of REPAIR relies on in-training classification loss as opposed to out-of-sample generalization accuracy. 
%[Related work] REPAIR \citep{Li2019REPAIRRR}. Contrasting motivation, different implementation.
RESOUND \citep{li2018resound} quantifies the representation biases of datasets, and uses them to assemble a new K-class dataset with smaller biases by sampling an existing C-class dataset ($C > K$). 
%Data selection methods such as dataset distillation \citep{Wang2018Dataset} aim to distill data to a few examples such that downstream performance is \textit{preserved}; by design, \papername{} is adversarial to high downstream performance.
%
Dataset distillation \citep{Wang2018Dataset} optimizes for a different objective function compared to \papername{}: it aims to synthesize a small number of instances to approximate the model trained on the original data. 
\citet{Dasgupta2018EvaluatingCI} introduce an NLI dataset that cannot be solved using only word-level knowledge and requires some compositionality. The authors show that debiasing training corpora and augmenting them with minimal contrasting examples makes models more suited to learn the compositional structure of language.
Finally, \citet{Sagawa2020AnIO} analyze the tension between over-parameterization and using all the data available. It advocates for subsampling the majority groups as opposed to upweighting minority groups in order to achieve low worst-group error. This is in line with the filtering approach that \papername{} adapts, as well as the out-of-distribution and robustness results we observe.

\paragraph{Learning Objectives for Debiasing} Another line of related work focuses on removing bias in data representations via the design of learning objectives for debiasing. 
\citet{Arjovsky2019InvariantRM} propose Invariant Risk Minimization as an objective that promotes learning representations of the data that are stable across environments. 
Instead of learning optimal classifiers, \papername{} aims to remove instances that exhibit artifacts in a dataset. 
\citet{Belinkov2019DontTT} propose an adversarial removal technique that encourages models to learn representations free of hypothesis-only biases.
\citet{He2019UnlearnDB} propose DRiFt, a debiasing algorithm that first learns a biased model using only known biased features and then trains a debiased model that fits the residuals of the biased model.
Similarly, \citet{clark-etal-2019-dont} propose learning a naive classifier using only bias features, to be used in an ensemble along with other classifiers containing more general features.
Each of the previous approaches target only \textit{known} NLI biases, based on prior knowledge; we show \papername{} is capable of removing even those examples which exhibit previously \textit{unknown} spurious biases. 
%\cite{Elazar2018AdversarialRO} show that demographic information leaks into
%intermediate representations of neural networks
%trained on text data.  
Finally, \citet{Elazar2018AdversarialRO} show that adversarial training effectively mitigate demographic information leakage, but fail to remove it completely when dealing with text data.

%We first learn a biased model
%that only uses features that are known to relate to dataset bias. Then, we train a debiased model that fits to the residual of the biased model, focusing on examples that cannot
%be predicted well by biased features only.

% argue that unstable, spurious correlations in the data would generalize poorly to novel test environments. 
% Thus, they 

%To add: 

%\cite{Rudinger2017SocialBI} found evidence that the elicited hypotheses introduced substantial gender stereotypes as well as varying degrees of racial, religious, and age-based stereotypes.

\section{Conclusion}
\label{sec:conclusion}

We present a deep-dive into \papername{} -- an iterative greedy algorithm that adversarially filters out spurious biases from data for accurate benchmark estimation. 
We provide a theoretical framework supporting \papername{}, and show its effectiveness in bias reduction on synthetic and real datasets, providing extensive analyses. 
We apply \papername{} to four datasets, including widely used benchmarks such as SNLI and ImageNet.  
On out-of-distribution and adversarial test sets designed for such benchmarks, we show that models trained on the \papername{}-filtered subsets achieve better performance, indicating higher generalization abilities.
Moreover, we show that the strongest performance on the resulting filtered datasets drops significantly (by 30 points for SNLI and 20 points for ImageNet).
We hope that dataset creators will employ \papername{} to identify unknown dataset artifacts before releasing new challenge datasets for more reliable estimates of task progress on future AI benchmarks. 
All datasets and code for this work will be made public.
\section*{Acknowledgments}
%We thank the anonymous reviewers for their insightful feedback.
We would like to thank Noah A. Smith, Nicholas Lourie, Ana Marasovi\`{c} and Daniel Khashabi for insightful discussions about this work as well as the anonymous reviewers for their valuable feedback.
This research was supported in part by NSF (IIS-1524371), the National Science Foundation Graduate Research Fellowship under Grant No. DGE 1256082, DARPA CwC through ARO (W911NF15-1- 0543), DARPA MCS program through NIWC Pacific (N66001-19-2-4031), and the Allen Institute for AI. 
Computations on \url{beaker.org} were supported in part by credits from Google Cloud.

% \subsubsection*{Acknowledgments}
\bibliography{references}
\bibliographystyle{icml2020}

\clearpage
\appendix
\section{Appendix}

%%%%%%%%%%%%%%%%%
\subsection{Filtering Heuristics}
\label{sec:heuristics}

We present three heuristic approaches that approximate the optimum bias reduction problem (\afopt):
\textbf{(A)} A simple \emph{greedy approach} starts with the full set $S = \mathcal{D}$, identifies an $i \in S$ that maximizes $\tilde{p}(i)$, removes it from $S$, and repeats up to $|\mathcal{D}|-n$ times.
\textbf{(B)} A \emph{greedy slicing approach} identifies the instances with the $k$ highest predictability scores, removes all of them from $S$, and repeats the process up to $\lfloor \frac{|\mathcal{D}|-n}{k} \rfloor$ times.
\textbf{(C)} A \emph{slice sampling approach}, instead of greedily choosing the top $k$ instances, randomly samples $k$ instances with probabilities proportional to their predictability scores (cf.~Appendix~\S\ref{sec:slice-sampling} for more details).
% The Gumbel method provides an efficient way to perform such sampling~\cite{Gumbel1954StatisticalTO,Maddison2014AS,Kim2016ExactSW,Balog2017LostRO,Kool2019StochasticBA}, by independently perturbing each $\tilde{p}(i)$ with a Gumbel random variable and identifying $k$ instances with the highest \emph{perturbed} predictability scores 

All three strategies could be further improved by considering not only the predictability score of the top-$k$ instances but also (via retraining without these instances) how their removal would influence the predictability scores of other instances in the next step. 
We found our computationally lighter approaches to work well even without the additional overhead of such look-ahead.
\papername{} implements the greedy slicing approach, and can thus be viewed as a scalable and practical approximation of (intractable) \afopt{} for optimum bias reduction.
We leave the empirical investigation into other proposed strategies for future work.

%%%%%%%%%%%%%%%%%
\subsection{Slice Sampling Details}
\label{sec:slice-sampling}

As discussed in Appendix \S\ref{sec:heuristics} (\textbf{C}), the \textit{slice sampling approach} can be efficiently implemented using what is known as the Gumbel method or Gumbel trick~\cite{Gumbel1954StatisticalTO,Maddison2014AS}, which uses random perturbations to turn sampling into a simpler problem of optimization. This has recently found success in several probabilistic inference applications~\cite{Kim2016ExactSW,Jang2016CategoricalRW,Maddison2016TheCD,Balog2017LostRO,Kool2019StochasticBA}. Starting with the log-predictability scores $\log \tilde{p}(i)$ for various $i$, the idea is to perturb them by adding an independent random noise $\gamma_i$ drawn from the standard Gumbel distribution. Interestingly, the maximizer $i^*$ of $\gamma_i + \log \tilde{p}(i)$ turns out to be an exact sample drawn from the (unnormalized) distribution defined by $\tilde{p}$. Note that $i^*$ is a random variable since the $\gamma_i$ are drawn at random. This result can be generalized~\citep{Vieira2014GumbelmaxTA} for slice sampling: the $k$ highest values of Gumbel-perturbed log-predictability scores correspond to sampling, without replacement, $k$ items from the probability distribution defined by $\tilde{p}$. The Gumbel method is typically applied to exponentially large combinatorial spaces, where it is challenging to scale up. In our setting, however, the overhead is minimal since the cost of drawing a random $\gamma_i$ is negligible compared to computing $\tilde{p}(i)$.

%%%% Replacing this table with a figure

\begin{table}
\centering
\small
\caption{Mean \emph{Dev} accuracy ($\%$) on two models trained on four synthetic datasets before ($D$) and after ($D(\Phi)$) \papername{}.
Standard deviation across 10 runs with randomly chosen seeds is provided as a subscript.
The datasets, also shown in Fig.~\ref{fig:synthetic} differ in the degree of separation between the two classes.
Both models (SVM with an RBF kernel \& linear classifier with logisitic regression) perform well on the original synthetic dataset, before filtering.
The linear classifier performs well on the data, because it contains spurious artifacts, making the task artificially easier for it.
However, after \papername{}, the linear model, relying mostly on the spurious features, clearly underperforms.
}
\begin{tabular}[t]{llcc}
\toprule
Class & & & \\
%               & \multicolumn{2}{c}{Synthetic}\\
%\cmidrule(lr){2-3}
Separation  &  Model            & $D$    & $D(\Phi)$ \\
\midrule
\multirow{2}{*}{0.8}  &  SVM-RBF & 97.0$_{02}$ & 90.7$_{06}$ \\
  &  Logistic Reg. & 83.5$_{05}$ & 50.7$_{12}$ \\ 
  \midrule[0.03em]
\multirow{2}{*}{0.7}  &  SVM-RBF & 89.9$_{04}$ & 82.5$_{09}$ \\
 &  Logistic Reg. & 74.3$_{11}$ & 52.4$_{13}$\\
 \midrule[0.03em]
\multirow{2}{*}{0.6}  &  SVM-RBF & 87.6$_{04}$& 77.8$_{09}$ \\
  &  Logistic Reg. & 74.3$_{10}$ & 53.1$_{12}$ \\ 
  \midrule[0.03em]
\multirow{2}{*}{0.4}  &  SVM-RBF & 83.8$_{05}$ & 70.7$_{10}$ \\
  &  Logistic Reg. & 75.4$_{09}$ & 53.4$_{12}$ \\
\bottomrule
\end{tabular}
\label{tab:synthetic_results}
\end{table}

\subsection{Results on Synthetic Data Experiments}
\label{sec:syn_results}

% \begin{figure*}[ht]
%     \centering
%     \includegraphics[width=\linewidth]{sections/figures/synthetic.png}
%     \caption{Four sample biased datasets as input to \papername (top). Blue and orange indicate two different classes. Only the original two dimensions are shown, not the bias features. For the leftmost dataset with the highest separation, we flip some labels at random, so even an RBF kernel cannot achieve perfect performance. \papername{} makes the data more challenging for the models (bottom).} 
%      \label{fig:synthetic_big}
% \end{figure*}

As discussed in Section \S\ref{sec:exp_syn}, Figure \ref{fig:synthetic} shows the effect of \papername on four synthetic datasets containing data arranged in concentric circles at four degrees of class separation.
% This adds two more experiments (shown on the extreme right) exhibiting similar phenomena as those shown in Figure \ref{fig:synthetic}.
For greater visibility, we have provided the accuracies of the SVM with RBF kernel and logistic regression in Table \ref{tab:synthetic_results}. 
% Results on the synthetic dataset are provided in Table \ref{tab:synthetic_results}. 
% Please refer to Section (\S\ref{sec:exp_syn}) for a detailed description.

In summary, a stronger model such as the SVM is more robust to the presence of artifacts than a simple linear classifier.
Thus, the implications for real datasets is to move towards models designed for reasoning about a specific task, hence avoiding a dependence on spurious artifacts.

\subsection{NLI Out-of-distribution Benchmarks}
\label{sec:nli-ood}

We describe the four out-of-distribution evaluation benchmarks for NLI from Section \S\ref{sec:snli-gen} below:
\begin{compactitem}
\item HANS \citep{mccoy-etal-2019-right} contains evaluation examples designed to avoid common structural heuristics (such as word overlap) which could be used by models to correctly predict NLI inputs, without true inferential reasoning.
\item NLI Diagnostics \citep{Wang2018GLUEAM} is a set of hand-crafted examples designed to demonstrate model performance on several fine-grained semantic categories, such as logical reasoning and commonsense knowledge.
\item Stress tests for NLI \cite{naik-etal-2018-stress} are a collection of tests targeting the weaknesses of strong NLI models, to check if these are robust to semantics (competence), irrelevance (distraction) and typos (noise).
\item Adversarial NLI \citep{Nie2019-ri} consists of premises collected from Wikipedia and other news corpora, and human generated hypotheses, arranged at different tiers of the challenge they present to a model, using a human and model in-the-loop procedure.
\end{compactitem}

Recent work \cite{McCoy2019BERTsOA} has observed large variance on out-of-distribution test sets with random seeds.
Hence, we report the mean and variance across 5 random seeds in all settings in Table \ref{tab:snli_gen}.
Since Adversarial NLI involves finetuning the model, and not just reporting on a different test set, we skip this step in Table \ref{tab:snli_adversarial}.

\subsection{Hyperparameters for \papername}
\label{sec:aflite_hyperparams}

\begin{table}
\small
\centering
\caption{Hyperparameters for the \papername algorithm, used for in-distribution benchmark estimation on different datasets.  
$m$ denotes the size of the support of the expectation in Eq.~(\ref{eq:predictability-score-approx}), $t$ is the training set size for the linear classifiers, $k$ is the size of each slice, and $\tau$ is an early-stopping filtering threshold. 
For ImageNet, we set $n=640K$ and hence do not need to control for $\tau$.
In every other setting, we set $\tau$ as above, and hence do not need to control for $n$.
Detailed definitions for each hyperparameter is provided in Section \S\ref{sec:algo}.
}
\begin{tabular}[t]{lrrrrr}
\toprule
  & Synthetic & SNLI & MultiNLI & QNLI  & ImageNet\\ 
\midrule
$m$         &  128 & 64 & 64 & 64 & 64\\
$t$         &  100 & 50K & 40K & 10K & 32.7K \\
$k$         &  1  & 10K & 10K & 2K & 33.6K \\
$\tau$      & 0.75 & 0.75 & 0.75 & 0.75 & - \\
\bottomrule
\end{tabular}
\label{tab:hyperparams_aflite}
\end{table}

Table \ref{tab:hyperparams_aflite} shows hyperparameters used to run \papername to obtain filtered subsets for in-distribution benchmark estimation on different datasets. 
Target dataset size, $n$ and the early stop filtering threshold $\tau$ are interdependent, as the predictability score threshold determines what examples to keep, which in turn influences the desired size of the dataset, $n$.
For ImageNet, we set $n=640K$ and do not control for $\tau$. 
We use much larger values for $t$ and $k$ for ImageNet than in all NLP experiments, where the use of powerful language representations (such as \roberta) allows us to get reasonable performance even with smaller training sets; ImageNet does not offer any such benefits arising from pretrained representations.

For all out-of-distribution NLP experiments, we explicitly control for the size of $n$, as discussed in the corresponding sections in the paper.
In these cases, we typically end up using slightly larger $n$, allowing for the final models to get more exposure to task data which is, to a degree, helpful for out-of-distribution generalization.
In ImageNet, we use the same hyperparameters in both sets of experiments.
In particular, we explicitly set $n=182K$ for SNLI, and $n=640K$ for ImageNet \papername-filtering for the out-of-distribution generalization experiments.

%The AFLite hyper-parameters, including the slice size (k) and the number of partitions (m), have a direct, intuitive interpretation in terms of their impact on the results. As suggested, we conducted additional experiments on SNLI, which show the robustness of AFLite w.r.t. k and m:

\subsection{Hyperparameters for NLP experiments}
\label{sec:nli_hyperparams}

For all NLP experiments, our implementation is based on the GLUE \cite{Wang2018GLUEAM} experiments in the Transformers repository \cite{Wolf2019HuggingFacesTS} from Huggingface.\footnote{\url{https://github.com/huggingface/transformers}}
We used the Adam optimizer \cite{kingma2014adam} for every training set up, with a learning rate of 1e-5, and an epsilon value of 1e-8.
We trained for 3 epochs for all *NLI tasks, maintaining a batch size of 92.
All above hyperparameters were selected using a grid search; we kept other hyperparameters unaltered from the original HuggingFace repository.
% Other hyperparameters are listed in Table \ref{tab:hyperparams}.
Each experiment was performed on a single Quadro RTX 8000 GPU.

\subsection{Hyperparameters for ImageNet}
\label{sec:imagenet_hyperparams}

We trained our ImageNet models using v3-512 TPU pods. 
For EfficientNet \cite{tan2019efficientnet}, we used RandAugment data augmentation \cite{cubuk2019randaugment} with 2 layers, and a magnitude of 28, for all model sizes. 
We trained our models using a batch size of 4096, a learning rate of 0.128, and kept other hyperparameters the same as in \cite{tan2019efficientnet}. 
We trained for 350 epochs for all dataset sizes - so when training on 20\% or 40\% of ImageNet (or a smaller dataset), we scaled the number of optimization steps accordingly. 
For ResNet \cite{he2016deep}, we used a learning rate of 0.1, a batch size of 8192, and trained for 90 epochs.

\subsection{Qualitative Analysis of SNLI}
\label{sec:qualitative_nli}

Table \ref{tab:examples} shows some examples removed and retained by \papername on the NLI dataset.

\begin{table*}[t]
\centering
\small
% \rotatebox{90}{
\caption{Examples from SNLI, removed (top) and retained (bottom) by \papername.
As is evident, the retained instances are slightly more challenging and capture more nuanced semantics in contrast to the removed instances.
Removed instances also exhibit larger word overlap, and many other artifacts found in \citet{gururangan-etal-2018-annotation}.
Two examples per label are shown, the \papername-filtered dataset contains many more \texttt{neutral} examples, as opposed to those labeled as \texttt{contradiction}.
}
\begin{tabular}[t]{p{7cm}p{6cm}r}
\toprule 

\multicolumn{3}{c}{\bf \textsc{Removed by \papername}} \\
 \midrule
\textbf{Premise}  & \bf Hypothesis & \bf Label\\
 \midrule[0.03em]
A woman, in a green shirt, preparing to run on a treadmill.   &  A woman is preparing to sleep on a treadmill. &  \texttt{contradiction} \\
The dog is catching a treat.  &  The cat is not catching a treat. &  \texttt{contradiction} \\
Three young men are watching a tennis match on a large screen outdoors. & Three young men watching a tennis match on a screen outdoors, because their brother is playing. & \texttt{\texttt{neutral}} \\
A girl dressed in a pink shirt, jeans, and flip-flops sitting down playing with a lollipop machine.  &   A funny person in a shirt. & \texttt{\texttt{neutral}} \\
A man in a green apron smiles behind a food stand.  &    A man smiles. &  \texttt{entailment} \\
A little girl with a hat sits between a woman's feet in the sand in front of a pair of colorful tents. & The girl is wearing a hat. & \texttt{entailment} \\
\midrule 
\multicolumn{3}{c}{\bf \textsc{Retained by \papername}} \\
 \midrule
\textbf{Premise}  & \bf Hypothesis & \bf Label\\
 \midrule[0.03em]
People are throwing tomatoes at each other.   &  The people are having a food fight. & \texttt{entailment} \\
 A man poses for a photo in front of a Chinese building by jumping.   &   The man is prepared for his photo. & \texttt{entailment} \\
An older gentleman speaking at a podium.    &    A man giving a speech & \texttt{\texttt{neutral}} \\
A man poses for a photo in front of a Chinese building by jumping.   &   The man has experience in taking photos. & \texttt{\texttt{neutral}} \\
 People are waiting in line by a food vendor.  &  People sit and wait for their orders at a nice sit down restaurant.  &   \texttt{contradiction} \\
Number 13 kicks a soccer ball towards the goal during children's soccer game. &  A player passing the ball in a soccer game. & \texttt{contradiction} \\
\bottomrule
\end{tabular}
% }
\label{tab:examples}
\end{table*}

\end{document}